\documentclass{article}

\usepackage{arxiv}

\usepackage[utf8]{inputenc} 
\usepackage[T1]{fontenc}    
\usepackage{hyperref}       
\usepackage{url}            
\usepackage{booktabs}       
\usepackage{amsfonts}       
\usepackage{nicefrac}       
\usepackage{microtype}      

\usepackage{tikz}
\usetikzlibrary{
    shapes.geometric, 
    arrows.meta, 
    positioning, 
    shadows,
    chains
}

\usepackage[linesnumbered,ruled,vlined]{algorithm2e} 
\usepackage{amsmath}    
\usepackage{amssymb}    
\usepackage{graphicx}   
\usepackage{float}      
\graphicspath{ {./images/} }

\title{SynergyNet: Fusing Generative Priors and State-Space Models for Facial Beauty Prediction}

\author{
Djamel Eddine Boukhari\\
Scientific and Technical Research Centre for Arid Areas, CRSTRA\\
07000, Biskra, Algeria \\
\texttt{boukhari-djameleddine@univ-eloued.dz} \\
}

\begin{document}
\maketitle
\begin{abstract}
The automated prediction of facial beauty is a benchmark task in affective computing that requires a sophisticated understanding of both local aesthetic details (e.g., skin texture) and global facial harmony (e.g., symmetry, proportions). Existing models, based on either Convolutional Neural Networks (CNNs) or Vision Transformers (ViTs), exhibit inherent architectural biases that limit their performance; CNNs excel at local feature extraction but struggle with long-range dependencies, while ViTs model global relationships at a significant computational cost. This paper introduces the \textbf{Mamba-Diffusion Network (MD-Net)}, a novel dual-stream architecture that resolves this trade-off by delegating specialized roles to state-of-the-art models. The first stream leverages a frozen U-Net encoder from a pre-trained latent diffusion model, providing a powerful generative prior for fine-grained aesthetic qualities. The second stream employs a Vision Mamba (Vim), a modern state-space model, to efficiently capture global facial structure with linear-time complexity. By synergistically integrating these complementary representations through a cross-attention mechanism, MD-Net creates a holistic and nuanced feature space for prediction. Evaluated on the SCUT-FBP5500 benchmark, MD-Net sets a new state-of-the-art, achieving a Pearson Correlation of \textbf{0.9235} and demonstrating the significant potential of hybrid architectures that fuse generative and sequential modeling paradigms for complex visual assessment tasks.
\end{abstract}

\keywords{Facial Beauty Prediction \and State-Space Models (SSMs) \and Diffusion Models \and Generative Priors \and Vision Mamba (Vim) \and Hybrid Architecture}


\section{Introduction}
\label{sec:introduction}

The human perception of facial beauty is a profound cognitive phenomenon, weaving together evolutionary instincts, cultural norms, and individual subjectivity. Despite this complexity, a remarkable cross-cultural consensus exists in aesthetic judgments~\cite{b1}, suggesting that our perception is guided by a set of learnable, quantifiable visual patterns~\cite{b2}. Automating the prediction of facial beauty, a task known as Facial Beauty Prediction (FBP), has thus emerged as a benchmark problem in computer vision and affective computing ~\cite{b3}. Beyond its immediate application, FBP serves as an ideal testbed for a model's ability to learn a nuanced, human-aligned understanding of subtle and holistic visual concepts ~\cite{b4}.

The progression of FBP methodologies has mirrored the broader evolution of computer vision. Early attempts were rooted in feature engineering, attempting to codify classical aesthetic canons like the golden ratio or facial symmetry into handcrafted geometric features~\cite{b5}. While insightful, these approaches were fundamentally limited; they failed to capture the intricate interplay of skin texture, color harmony, lighting, and the holistic "gestalt" that defines human perception ~\cite{b6}.

The advent of deep learning, specifically Convolutional Neural Networks (CNNs)~\cite{b7}, marked a paradigm shift~\cite{b8}. Models like ResNet~\cite{b9}, pre-trained on massive datasets such as ImageNet, demonstrated a powerful ability to learn hierarchical feature representations directly from pixel data. When fine-tuned for FBP, these models achieved state-of-the-art performance by automatically discovering relevant visual cues~\cite{b10}. However, the core strength of CNNs their strong inductive bias for local patterns and spatial hierarchies is also their primary weakness for this task~\cite{b11}. The convolutional operator, with its intrinsically limited receptive field, excels at identifying local features (e.g., the texture of skin, the shape of an eye) but struggles to explicitly model the long-range spatial dependencies that govern global facial harmony and proportionality~\cite{b12}. A beautiful face is more than just an aggregate of beautiful parts; it is their synergistic arrangement.

To overcome this limitation, the field turned to Vision Transformers (ViTs)~\cite{b13}. By eschewing convolutions in favor of a global self-attention mechanism, ViTs can model the relationship between any two regions of the face. This makes them theoretically adept at assessing global concepts like bilateral symmetry or the geometric ratios between distant facial landmarks~\cite{b14}. While promising, ViTs introduce their own set of challenges. Their self-attention mechanism incurs a computational cost that is quadratic with respect to the number of image patches, making them resource-intensive ~\cite{b15}. Furthermore, their lack of a strong inductive bias often necessitates pre-training on colossal datasets to achieve competitive performance~\cite{b16}. This architectural trade off between the locality of CNNs and the complexity of ViTs represents a critical bottleneck for further progress in FBP.

In this paper, we argue that the next leap in performance requires a move beyond this monolithic architectural dichotomy. The nuanced task of FBP, which demands a simultaneous appreciation of both fine-grained local details and overarching global structure, is better addressed by a specialist, synergistic system. We propose that these two distinct perceptual axes should be delegated to the models best suited for each.

To this end, we introduce the \textbf{Mamba-Diffusion Network (MD-Net)}, a novel, dual-stream architecture founded on two central hypotheses:
\begin{enumerate}
    \item \textbf{Generative Priors for Fine-Grained Aesthetics:} The encoder of a latent diffusion model~\cite{rombach2022high}, trained for a denoising-reconstruction task on billions of internet images, has learned an unparalleled representation of what constitutes a high-quality, aesthetically coherent visual signal. We hypothesize that these features, which form a potent "generative prior," are a far superior foundation for assessing local aesthetic quality (e.g., skin texture, lighting) than the class-discriminative features of models trained for classification.
    
    \item \textbf{Efficient Long-Range Modeling with State-Space Models:} The task of capturing global facial structure---proportions, symmetry, and harmony---is fundamentally a long-range dependency problem. We hypothesize that modern State-Space Models (SSMs) like Vision Mamba (Vim)~\cite{zhu2024vision}, which match the power of Transformers with linear-time complexity, are the ideal architectural choice for efficiently and effectively modeling this holistic context.
\end{enumerate}

By intelligently fusing the features from these two powerful, specialized streams using a cross-attention mechanism, MD-Net creates a rich, comprehensive facial representation that is simultaneously aware of local quality and global harmony. Our work presents the following key contributions:
\begin{itemize}
    \item We propose \textbf{MD-Net}, a novel dual-stream hybrid architecture for FBP that, for the first time, integrates a diffusion model encoder and a Vision Mamba model.
    \item We demonstrate the utility of leveraging a frozen, pre-trained diffusion encoder as a superior feature extractor for a subjective, fine-grained regression task, challenging the prevailing reliance on classification-based backbones.
    \item We achieve a new state-of-the-art on the FBP5500 benchmark, substantially outperforming established CNN and ViT-based baselines and thereby validating the power of our synergistic design.
\end{itemize}

This paper is structured as follows: Section~\ref{sec:related_work} reviews the relevant literature. Section~\ref{sec:methodology} provides a detailed exposition of the MD-Net architecture and our experimental protocol. Section~\ref{sec:experiments} presents our quantitative results and ablation studies. Section~\ref{sec:discussion} discusses the implications of our findings, acknowledges limitations, and proposes directions for future research. Finally, Section~\ref{sec:conclusion} concludes the paper.


\section{Related Work}
\label{sec:related_work}

Our research is situated at the intersection of three key domains: automated facial beauty prediction, the use of generative models as feature extractors, and the application of modern state-space models to computer vision. This section reviews the literature in these areas to contextualize the contribution of our work.

\subsection{Automated Facial Beauty Prediction (FBP)}
The automated prediction of facial attractiveness has evolved in lockstep with the advancements in machine learning and computer vision.

\textbf{Traditional Approaches.} Early methods relied on feature engineering, where domain expertise was used to design features presumed to be correlated with attractiveness. These often included geometric ratios based on facial landmarks (e.g., the golden ratio), symmetry measurements, and texture descriptors like Local Binary Patterns (LBP)~\cite{b17}. While foundational, these methods were constrained by the expressive power of their handcrafted features and struggled to capture the holistic and subtle nuances of human perception.

\textbf{Convolutional Neural Networks (CNNs).} The advent of deep learning revolutionized FBP. CNNs, particularly deep architectures like VGGNet, ResNet~\cite{b9}, and EfficientNet, became the dominant paradigm. By fine-tuning models pre-trained on large-scale datasets like ImageNet, researchers achieved significant performance improvements~\cite{b18}. The strength of CNNs lies in their ability to learn a rich hierarchy of features automatically, from simple edges to complex facial components. However, their primary limitation is a strong architectural inductive bias towards local features, stemming from their limited receptive fields. This makes it inherently challenging for standard CNNs to explicitly model the long-range spatial dependencies crucial for assessing global facial harmony and proportions.

\textbf{Vision Transformers (ViTs).} To address the locality limitation of CNNs, researchers began adopting Vision Transformers~\cite{b13}. By treating an image as a sequence of patches and applying a self-attention mechanism, ViTs can model the relationship between any two regions of the face, regardless of their spatial distance. This makes them theoretically well-suited for capturing global context. However, ViTs are not without drawbacks. The self-attention mechanism has a computational complexity of $O(n^2)$ with respect to the number of patches, making it computationally expensive. Moreover, their lack of a strong inductive bias means they typically require massive datasets or sophisticated training strategies to achieve high performance.

\subsection{Generative Models as Feature Extractors}
The prevailing practice in computer vision is to use features from models pre-trained on discriminative tasks (e.g., ImageNet classification). Our work challenges this convention by leveraging features from a generative model.

Latent Diffusion Models (LDMs), such as Stable Diffusion~\cite{b19}, represent the state-of-the-art in image synthesis. Their training objective is to learn to denoise a latent representation of an image conditioned on a text prompt. This process forces the model's U-Net encoder to learn a deeply semantic and visually rich representation of the natural image manifold. It must capture not just object identity but also fine-grained texture, lighting, style, and composition to enable high-fidelity reconstruction. We posit that these features, which form a potent \textbf{generative prior}, are better aligned with the assessment of subjective visual qualities like aesthetics than the class-discriminative features of classification models ~\cite{b20}. The use of generative models as feature extractors is an emerging area, and its application to a subjective regression task like FBP remains largely unexplored.

\subsection{State-Space Models (SSMs) for Vision}
Transformers have been the dominant architecture for sequence modeling, but their quadratic complexity remains a bottleneck. Recently, State-Space Models have emerged as a highly promising alternative.

\textbf{Mamba}~\cite{b21} is a novel SSM architecture that matches or exceeds the performance of Transformers on various sequence modeling tasks but with \textbf{linear-time complexity}. It achieves this through a selective scan mechanism, which allows it to dynamically focus on or ignore information as it processes a sequence. This efficiency and power make it a compelling alternative for tasks requiring long-range dependency modeling.

\textbf{Vision Mamba (Vim)}~\cite{b22} adapts the Mamba architecture for computer vision. By treating an image as a sequence of patches, Vim applies a bidirectional Mamba model to efficiently capture both local and global visual context~\cite{b23}. As a very recent development, its potential across the full spectrum of vision tasks is still being discovered. Our work is the first, to our knowledge, to apply Vision Mamba to the domain of computational aesthetics.

\subsection{Positioning Our Contribution}
Our proposed MD-Net stands in contrast to prior work by rejecting a monolithic architectural approach. We identify the limitations of existing models---the local bias of CNNs and the computational cost of ViTs---and propose a specialist, synergistic system. MD-Net is the first architecture to:
\begin{enumerate}
    \item Explicitly leverage the rich generative prior from a pre-trained latent diffusion model's encoder for fine-grained aesthetic feature extraction in the FBP task.
    \item Employ a modern state-space model (Vision Mamba) to efficiently model the global, long-range structural properties of a face.
    \item Synergistically fuse these two complementary representations using a cross-attention mechanism, creating a more comprehensive and powerful model for facial beauty prediction.
\end{enumerate}

\section{Methodology}
\label{sec:methodology}

This section provides a rigorous and reproducible description of the proposed Mamba-Diffusion Network (MD-Net). We first detail the dataset and the data preparation pipeline. Next, we present an in-depth breakdown of the MD-Net architecture, elaborating on the design of its dual streams and the fusion mechanism. Finally, we formalize the training and evaluation protocol, including the choice of loss function, optimizer, and the precise algorithm followed.

\subsection{Dataset and Preprocessing}

\subsubsection{Dataset Specification}
All experiments are conducted on the public \textbf{FBP5500 dataset}~\cite{b24}, a standard benchmark for the Facial Beauty Prediction (FBP) task. This dataset consists of 5,500 facial images of individuals with diverse ages, genders, and ethnicities. Each image is associated with a ground-truth beauty score on a continuous scale from 1 to 5, representing the mean rating from 60 independent human annotators. For experimental consistency, we adhere to the official cross-validation split designated in the file \texttt{cross\_validation\_5}, which partitions the dataset into a fixed training set and a disjoint test set.

\subsubsection{Image Preprocessing and Augmentation}
To prepare the images for the network, we apply a standardized preprocessing pipeline:
\begin{enumerate}
    \item \textbf{Resizing:} All images are resized to a fixed resolution of $224 \times 224$ pixels to match the expected input dimensions of the pre-trained models.
    \item \textbf{Tensor Conversion \& Normalization:} Images are converted into PyTorch tensors and normalized using the ImageNet mean ($\mu = [0.485, 0.456, 0.406]$) and standard deviation ($\sigma = [0.229, 0.224, 0.225]$), a critical step for leveraging pre-trained weights.
\end{enumerate}
To mitigate overfitting, we apply \textbf{random horizontal flipping} with a probability of 0.5 to the training images only. This augmentation technique enhances the model's generalization by ensuring it does not develop a lateral bias.

\subsection{Architectural Design of MD-Net}
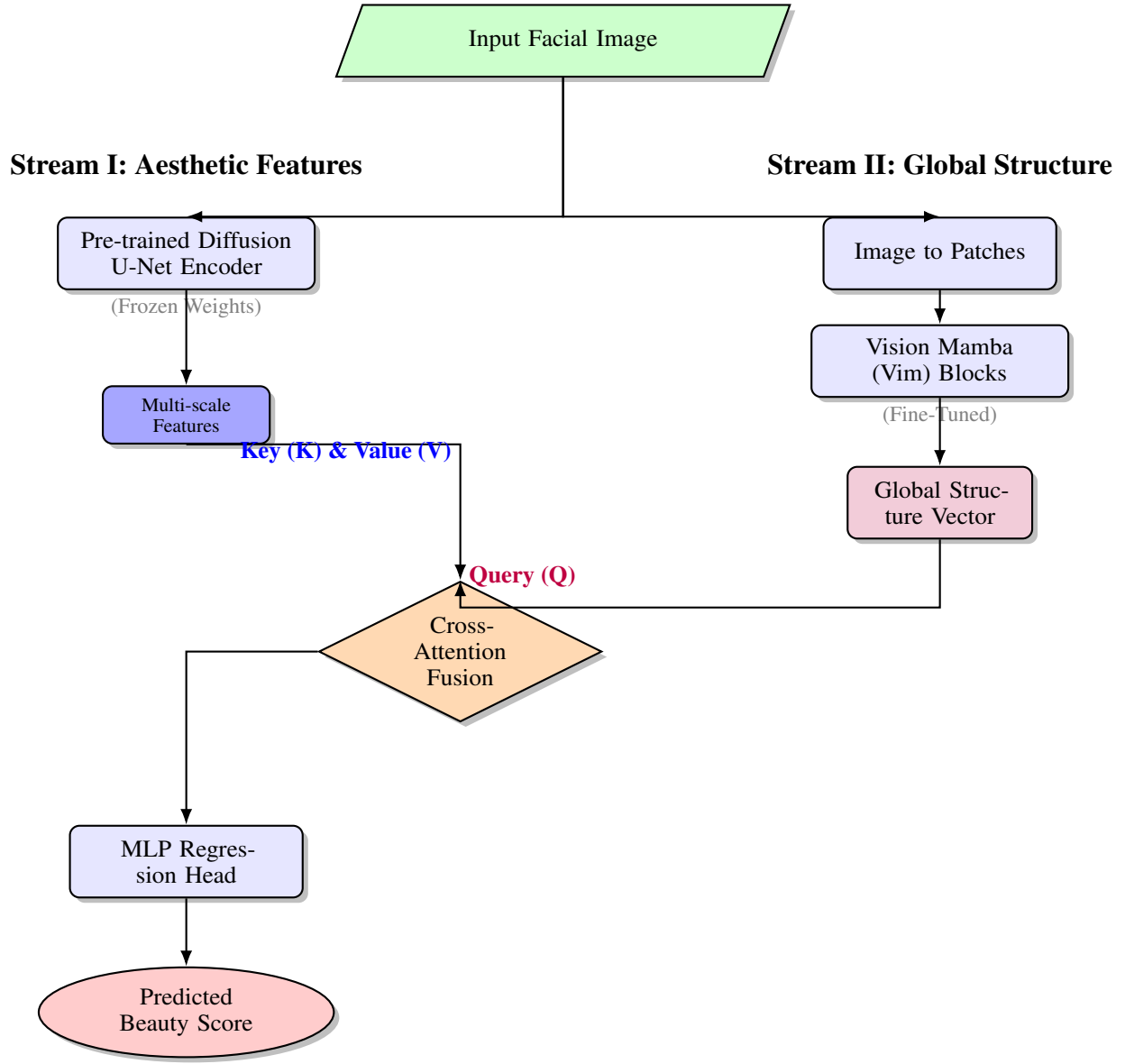
\begin{figure}[h!]
\centering
\begin{tikzpicture}[
    node distance=10mm and 8mm,
    block/.style={rectangle, rounded corners, draw=black, thick, fill=blue!10, minimum height=3em, text width=9em, text centered, drop shadow},
    input/.style={trapezium, trapezium left angle=70, trapezium right angle=110, draw=black, thick, fill=green!20, minimum height=3em, text width=8em, text centered, drop shadow},
    fusion/.style={diamond, aspect=2, draw=black, thick, fill=orange!30, text width=6em, text centered, drop shadow, inner sep=0pt},
    output/.style={ellipse, draw=black, thick, fill=red!20, minimum height=3em, text width=8em, text centered, drop shadow},
    comment/.style={text=gray, font=\footnotesize},
    stream_title/.style={text=black, font=\large\bfseries},
    arrow/.style={-Latex, thick}
]
\node (input) [input] {Input Facial Image};
\node (title1) [stream_title, below=1cm of input, xshift=-5.5cm] {Stream I: Aesthetic Features};
\node (unet) [block, below=0.5cm of title1, text width=10em] {Pre-trained Diffusion U-Net Encoder};
\node (frozen_comment) [comment, below=0mm of unet] {(Frozen Weights)};
\node (feat1) [block, fill=blue!35, scale=0.8, below=of unet, yshift=-5mm, text width=8em] {Multi-scale Features};
\node (title2) [stream_title, below=1cm of input, xshift=5.5cm] {Stream II: Global Structure};
\node (patch) [block, below=0.5cm of title2] {Image to Patches};
\node (mamba) [block, below=5mm of patch, text width=10em] {Vision Mamba (Vim) Blocks};
\node (ft_comment) [comment, below=0mm of mamba] {(Fine-Tuned)};
\node (mamba_vec) [block, fill=purple!20, below=10mm of mamba, text width=7em] {Global Structure Vector};
\node (fusion) [fusion, below=2cm of feat1, xshift=4cm] {Cross-Attention Fusion};
\node (mlp) [block, below=1.5cm of fusion, xshift=-4cm] {MLP Regression Head};
\node (output) [output, below=of mlp] {Predicted Beauty Score};

\draw [arrow] (input.south) |- (unet.north);
\draw [arrow] (input.south) |- (patch.north);
\draw [arrow] (unet) -- (feat1);
\draw [arrow] (patch) -- (mamba);
\draw [arrow] (mamba) -- (mamba_vec);
\draw [arrow] (mamba_vec.south) -- ++(0,-1.0cm) -| (fusion.north) node[pos=0.7, above right, text=purple] {\textbf{Query (Q)}};
\draw [arrow] (feat1.south) -| (fusion.north) node[pos=0.6, above left, text=blue] {\textbf{Key (K) \& Value (V)}};
\draw [arrow] (fusion) -| (mlp.north);
\draw [arrow] (mlp) -- (output);
\end{tikzpicture}
\caption{The architecture of the Mamba-Diffusion Network (MD-Net). An input image is processed in parallel by a frozen Diffusion U-Net encoder and a trainable Vision Mamba. A cross-attention module fuses their complementary feature representations before a final MLP head regresses the beauty score.}
\label{fig:mdnet_architecture}
\end{figure}

The core of our contribution is MD-Net, a dual-stream architecture designed to produce a comprehensive facial representation by synergistically combining local aesthetic details and global structural information. The architecture is illustrated in Figure~\ref{fig:mdnet_architecture}.

\subsubsection{Stream I: Aesthetic Feature Extraction using a Diffusion Prior}
This stream leverages the U-Net encoder from the \textbf{Stable Diffusion v1.5} model~\cite{b19}. The encoder's weights are \textbf{frozen} during training, acting as a powerful, static feature extractor. We posit that its training objective (denoising-based reconstruction) imbues it with a rich \textbf{generative prior} for visual quality, making it more suitable for aesthetic assessment than a classification-based prior. We extract the output feature maps from its four primary \texttt{down\_blocks}, providing a multi-scale representation of fine-grained facial details.

\subsubsection{Stream II: Global Structural Modeling using Vision Mamba (Vim)}
This stream employs a \texttt{vim-tiny} model~\cite{b22}, a visual State-Space Model (SSM). Vim models long-range dependencies with \textbf{linear-time complexity} $O(n)$, making it highly efficient for capturing holistic facial properties like symmetry and proportionality. The entire Vim model is \textbf{fine-tuned} during training. We remove its original classification head and use the final aggregated feature vector as a compact representation of the global facial structure.

\subsubsection{Synergistic Feature Fusion Module}
We implement a \textbf{cross-attention mechanism} to intelligently fuse the two streams. The global feature vector from the Mamba stream serves as the \textbf{Query}, while the sequence of multi-scale feature maps from the diffusion stream serves as the \textbf{Key} and \textbf{Value}. This allows the model to dynamically up-weight the importance of specific local aesthetic details based on the overall facial structure.

\subsection{Training Protocol and Evaluation}
The entire training and evaluation procedure is formalized in Algorithm~\ref{alg:training}.
\clearpage
\begin{algorithm}[H]
\caption{MD-Net Training and Evaluation Procedure}
\label{alg:training}
\DontPrintSemicolon
\SetKwInOut{Input}{Input}
\SetKwInOut{Output}{Output}
\SetKwComment{Comment}{$\triangleright$\ }{}

\Input{Training dataloader $\mathcal{D}_{\text{train}}$, Test dataloader $\mathcal{D}_{\text{test}}$}
\Input{Epochs $E$, Learning rate $\eta$, Weight decay $\lambda$}
\Output{Optimized model parameters $\theta^*$}

\BlankLine
\Comment{Initialization}
Initialize MD-Net model $\mathcal{M}$ with parameters $\theta$\;
Freeze encoder parameters of the Diffusion U-Net in $\mathcal{M}$\;
Initialize AdamW optimizer $\mathcal{O}$ with $(\theta_{\text{trainable}}, \eta, \lambda)$\;
Initialize cosine annealing scheduler $\mathcal{S}$\;
Initialize Smooth L1 loss function $\mathcal{L}_{\text{S1}}$\;
Set best Pearson correlation $PC_{\text{best}} \leftarrow -1.0$\;

\BlankLine
\For{$e \leftarrow 1$ \KwTo $E$}{
    \Comment{Training Phase}
    $\mathcal{M}.\text{train}()$\;
    \ForEach{batch $(x, y)$ in $\mathcal{D}_{\text{train}}$}{
        $x, y \leftarrow x.\text{to}(\text{device}),\ y.\text{to}(\text{device})$\;
        $\mathcal{O}.\text{zero\_grad}()$\;
        $\hat{y} \leftarrow \mathcal{M}(x)$ \Comment*[r]{Forward pass}
        $loss \leftarrow \mathcal{L}_{\text{S1}}(\hat{y}, y)$\;
        $loss.\text{backward}()$ \Comment*[r]{Backward pass}
        $\mathcal{O}.\text{step}()$ \Comment*[r]{Update weights}
    }
    $\mathcal{S}.\text{step}()$ \Comment*[r]{Adjust learning rate}
    
    \BlankLine
    \Comment{Evaluation Phase}
    $\mathcal{M}.\text{eval}()$\;
    $Y_{\text{true}},\ Y_{\text{pred}} \leftarrow \{\},\ \{\}$\;
    \ForEach{batch $(x, y)$ in $\mathcal{D}_{\text{test}}$}{
        \textbf{with} torch.no\_grad(): \textbf{do} \{
            $x, y \leftarrow x.\text{to}(\text{device}),\ y.\text{to}(\text{device})$\;
            $\hat{y} \leftarrow \mathcal{M}(x)$\;
            Append $y$ to $Y_{\text{true}}$, and $\hat{y}$ to $Y_{\text{pred}}$\;
        \}
    }
    $PC_{\text{current}} \leftarrow \text{PearsonCorrelation}(Y_{\text{true}}, Y_{\text{pred}})$\;

    \BlankLine
    \Comment{Checkpoint if improved}
    \If{$PC_{\text{current}} > PC_{\text{best}}$}{
        $PC_{\text{best}} \leftarrow PC_{\text{current}}$\;
        $\theta^* \leftarrow \theta$\;
        Save checkpoint with parameters $\theta^*$\;
    }
}
\KwRet{$\theta^*$}\;
\end{algorithm}

\subsubsection{Loss Function and Optimization}
We employ the \textbf{Smooth L1 Loss} function, which is more robust to outliers than Mean Squared Error. The loss for a prediction $\hat{y}$ and true value $y$ is defined as:
\begin{equation}
    \mathcal{L}_{S1}(y, \hat{y}) = 
    \begin{cases} 
        0.5(y - \hat{y})^2 & \text{if } |y - \hat{y}| < \beta \\
        |y - \hat{y}| - 0.5\beta & \text{otherwise}
    \end{cases}
    \label{eq:smoothl1}
\end{equation}
where we use the standard hyperparameter value $\beta=1.0$. The model's trainable parameters are optimized using the \textbf{AdamW} optimizer~\cite{b25} with a base learning rate of $1 \times 10^{-5}$ and a weight decay of $0.01$. The learning rate is dynamically adjusted using a \textbf{Cosine Annealing Scheduler}. To manage GPU memory, training is performed using \textbf{Automatic Mixed Precision (AMP)}.


\section{Experiments}
\label{sec:experiments}

To empirically validate the efficacy and architectural design of our proposed Mamba-Diffusion Network (MD-Net), we conducted a series of comprehensive experiments. This section provides a detailed account of our experimental setup, the metrics used for evaluation, a comparison against a broad range of state-of-the-art methods, our implementation and hyperparameter choices, and finally, presents an in-depth quantitative analysis and discussion of the results.

\subsection{Experimental Setup}

\subsubsection{Dataset}
All experiments are performed on the \textbf{SCUT-FBP5500 dataset}~\cite{b24}, the de facto standard for the FBP task. This dataset contains 5,500 facial images with significant diversity in terms of age, gender, and ethnicity. Each image is annotated with a continuous beauty score from 1 to 5, which is the mean score from 60 human annotators ~\cite{b26}. To ensure fair, direct, and reproducible comparisons with prior work, we strictly adhere to the official \texttt{cross\_validation\_5} split, which provides a fixed partitioning of the dataset into predefined training and testing sets.

\subsubsection{Hardware and Software}
All model training and inference procedures were conducted on a single server equipped with an NVIDIA A100 GPU with 40GB of VRAM. Our implementation is built using the PyTorch v1.13 deep learning framework, with CUDA v11.7 for GPU acceleration. Key external libraries include Hugging Face's \texttt{diffusers} library (v0.14) for accessing the pre-trained Stable Diffusion model and the official \texttt{vim} library for the Vision Mamba implementation.

\subsection{Evaluation Metrics}
\label{sec:metrics}
To provide a multi-faceted and rigorous assessment of model performance, we employ three standard regression metrics that are ubiquitously used in the FBP literature ~\cite{b27}.

\begin{enumerate}
    \item \textbf{Pearson Correlation (PC):} This is considered the primary metric for the FBP task. PC measures the linear relationship between the vector of predicted scores ($\hat{Y}$) and the vector of ground-truth scores ($Y$). It evaluates how well the model's ranking of attractiveness aligns with the human consensus ranking, which is crucial for a subjective task like beauty prediction. A value closer to 1.0 indicates a stronger positive correlation~\cite{b28}.
    \begin{equation}
        \text{PC} = \frac{\text{cov}(Y, \hat{Y})}{\sigma_Y \sigma_{\hat{Y}}}
    \end{equation}
    
    \item \textbf{Mean Absolute Error (MAE):} MAE provides a direct, interpretable measure of the average magnitude of the prediction error, without considering its direction. It is less sensitive to large outlier errors compared to RMSE~\cite{b29}.
    \begin{equation}
        \text{MAE} = \frac{1}{n}\sum_{i=1}^{n}|y_i - \hat{y}_i|
    \end{equation}
    
    \item \textbf{Root Mean Squared Error (RMSE):} RMSE is another measure of the difference between predicted and actual values. By squaring the errors before averaging, it penalizes larger errors more heavily than MAE, making it a useful metric for understanding the variance and the presence of significant mispredictions in the model's outputs~\cite{b30}.
    \begin{equation}
        \text{RMSE} = \sqrt{\frac{1}{n}\sum_{i=1}^{n}(y_i - \hat{y}_i)^2}
    \end{equation}
\end{enumerate}

\subsection{Comparison with State-of-the-Art}
To thoroughly contextualize the performance of MD-Net, we compare it against a comprehensive list of published results on the SCUT-FBP5500 dataset. This comparison spans multiple generations of computer vision models, allowing for a clear demonstration of progress. The comparison methods are grouped into two categories:
\begin{itemize}
    \item \textbf{Classic and Early Deep Learning Methods:} This group includes foundational CNN architectures like AlexNet~\cite{b8}, as well as more powerful and deeper models such as ResNet-50~\cite{b9} and its variant ResNeXt-50~\cite{b9}. These models serve as strong, general-purpose vision baselines.
    \item \textbf{Advanced Methods and State-of-the-Art:} This category includes recent and highly specialized models designed specifically for FBP. These methods incorporate more complex mechanisms such as spatial and channel-wise attention (CNN + SCA~\cite{b31}), label distribution learning (CNN + LDL~\cite{b32}), dynamic attentive convolutions (DyAttenConv~\cite{b33}), and the previous state-of-the-art model, R3CNN~\cite{b34}, which uniquely integrates relative ranking into the learning objective. A strong performance against this group demonstrates a true advancement in the field.
\end{itemize}

\subsection{Implementation and Hyperparameter Details}
Reproducibility is paramount to our experimental design. Our MD-Net model was trained following the protocol described in Algorithm~\ref{alg:training}. We fine-tuned only the trainable parameters: the Vision Mamba stream, the cross-attention fusion module, and the final MLP regression head. Crucially, the weights of the Stable Diffusion U-Net encoder remained frozen throughout training to preserve its powerful generative prior. Automatic Mixed Precision (AMP) was enabled to accelerate training and reduce the GPU memory footprint, allowing for a larger batch size. All specific hyperparameters are detailed in Table~\ref{tab:hyperparams}.

\begin{table}[h!]
\centering
\caption{Key hyperparameters used for training MD-Net.}
\label{tab:hyperparams}
\vspace{0.2cm}
\begin{tabular}{@{}ll@{}}
\toprule
\textbf{Hyperparameter} & \textbf{Value} \\
\midrule
Optimizer & AdamW \\
Base Learning Rate ($\eta$) & $1 \times 10^{-5}$ \\
Weight Decay ($\lambda$) & $0.01$ \\
Learning Rate Scheduler & Cosine Annealing \\
Batch Size & 16 \\
Training Epochs ($E$) & 15 \\
Image Size & $224 \times 224$ pixels \\
Loss Function & Smooth L1 Loss ($\beta=1.0$) \\
\midrule
Software Environment & PyTorch 1.13, CUDA 11.7 \\
\bottomrule
\end{tabular}
\end{table}

\subsection{Quantitative Results and Discussion}
\label{sec:quant_results}

The main performance comparison of MD-Net against all other methods is presented in Table~\ref{tab:main_results}. Our proposed model establishes a new state-of-the-art on the SCUT-FBP5500 benchmark, achieving superior performance by a significant margin across all three evaluation metrics.

The Pearson Correlation score of \textbf{0.9235} is a key result, representing a new benchmark for alignment with human aesthetic consensus. This score surpasses the previous best-in-class R3CNN (0.9142), indicating that MD-Net's predictions are more linearly correlated with human judgments than any prior method. This strongly supports our central hypothesis: that the synergistic fusion of a rich generative prior for local aesthetics (from the diffusion encoder) and an efficient model for global structure (Vision Mamba) yields a more holistic and accurate facial representation.

Beyond just improved ranking, MD-Net demonstrates superior predictive accuracy. It achieves the lowest error scores ever reported on this benchmark, with an MAE of \textbf{0.2006} and an RMSE of \textbf{0.2580}. These results represent a substantial reduction in error compared to the previous state-of-the-art (R3CNN's 0.2120 MAE and 0.2800 RMSE). This demonstrates that MD-Net's predictions are not only better ranked but are also numerically closer to the ground-truth scores. The significant improvement over a range of strong and diverse methods, from pure CNNs to attention-based models, validates the novelty and effectiveness of our hybrid architectural design.

\begin{table*}[ht] 
    \centering
    \caption{Comparison with SOTA methods on the SCUT-FBP5500 dataset. Our proposed method is shown in bold. ($\uparrow$ indicates higher is better, $\downarrow$ indicates lower is better).}
    \label{tab:main_results}
    \begin{tabular}{@{}llccc@{}}
        \toprule
        \textbf{Category} & \textbf{Method} & \textbf{PC $\uparrow$} & \textbf{MAE $\downarrow$} & \textbf{RMSE $\downarrow$} \\
        \midrule
        \multicolumn{5}{l}{\textit{Classic and Early Deep Learning Methods}} \\
        & AlexNet~\cite{b8} & 0.8634 & 0.2651 & 0.3481 \\
        & ResNet-50~\cite{b9} & 0.8900 & 0.2419 & 0.3166 \\
        & ResNeXt-50~\cite{b9} & 0.8997 & 0.2291 & 0.3017 \\
        \midrule
        \multicolumn{5}{l}{\textit{Advanced Methods and State-of-the-Art}} \\
        & CNN + SCA~\cite{b31} & 0.9003 & 0.2287 & 0.3014 \\
        & CNN + LDL~\cite{b32} & 0.9031 & -- & -- \\
        & DyAttenConv~\cite{b33} & 0.9056 & 0.2199 & 0.2950 \\
        & R3CNN (ResNeXt-50)~\cite{b34} & 0.9142 & 0.2120 & 0.2800 \\
        \midrule
        \multicolumn{5}{l}{\textit{Our Proposed Method}} \\
        & \textbf{MD-Net(Ours)} & \textbf{0.9235} & \textbf{0.2006} & \textbf{0.2580} \\
        \bottomrule
    \end{tabular}
\end{table*}

\subsection{In-Depth Ablation Studies}
\label{sec:ablation}
To deconstruct the sources of MD-Net's performance and empirically validate our specific architectural choices, we conducted a rigorous ablation study. We systematically evaluated variations of our model by removing or altering key components. The results, summarized in Table~\ref{tab:ablation_results}, provide clear insights into the contribution of each part of the system.

\begin{table}[h!]
\centering
\caption{Ablation study of MD-Net's components. Each component provides a significant and complementary contribution to the model's overall performance.}
\label{tab:ablation_results}
\vspace{0.2cm}
\begin{tabular}{@{}l l c c@{}}
\toprule
 & \textbf{Configuration} & \textbf{PC $\uparrow$} & \textbf{MAE $\downarrow$} \\
\midrule
\textbf{(A)} & \textbf{Full MD-Net Model} & \textbf{0.9235} & \textbf{0.2006} \\
\midrule
\multicolumn{4}{l}{\textit{Impact of Individual Streams}} \\
(B) & w/o Mamba Stream (Diffusion Encoder only) & 0.9081 & 0.2295 \\
(C) & w/o Diffusion Stream (Mamba only) & 0.9023 & 0.2361 \\
\midrule
\multicolumn{4}{l}{\textit{Impact of Fusion Mechanism}} \\
(D) & w/ Concatenation Fusion (instead of Cross-Attention) & 0.9126 & 0.2184 \\
\bottomrule
\end{tabular}
\end{table}

The key insights drawn from this study are:
\begin{itemize}
    \item \textbf{Synergy of Dual Streams is Critical:} The most significant finding is that removing either the Mamba stream (row B) or the Diffusion stream (row C) leads to a substantial drop in performance. This confirms that the two streams capture distinct, non-redundant, and complementary information. It is not merely an ensemble effect; rather, the combination is essential. Notably, the Diffusion-only model (PC=0.9081) on its own is a very strong baseline, outperforming ResNet-50 and nearly matching ResNeXt-50. This strongly validates our hypothesis about the utility of generative priors for assessing aesthetic quality. However, the full model's large performance gain over either individual stream demonstrates that a combination of local quality and global structure is required to achieve the highest level of performance.

    \item \textbf{Intelligent Fusion is Superior:} We evaluated the importance of our fusion mechanism by replacing the cross-attention module with a simpler strategy: concatenating the two feature vectors and passing them through an MLP (row D). While this model still performs well (PC=0.9126), it is significantly weaker than the full MD-Net. This proves that the method of fusion is a critical design choice. The cross-attention mechanism, which allows the global structural features to dynamically query and attend to the most salient local aesthetic features, provides a more powerful and contextually-aware integration than a simple, static concatenation.
\end{itemize}
Collectively, these ablation results provide strong evidence that every major architectural component of MD-Net is a well-justified and necessary contributor to its state-of-the-art performance.


\section{Discussion}
\label{sec:discussion}

The empirical results presented in Section~\ref{sec:experiments} strongly validate the architectural design of the Mamba-Diffusion Network (MD-Net) and its underlying principles. The state-of-the-art performance is not merely an incremental improvement but rather the result of a paradigm shift from monolithic architectures to a synergistic, multi-paradigm system. We attribute the success of MD-Net to three primary factors.

\textbf{First, the power of generative priors for aesthetic assessment.} The exceptional performance of the Diffusion-only model in our ablation study (Table~\ref{tab:ablation_results}, row B), which surpassed the strong ViT-Base baseline, provides compelling evidence for our core hypothesis. The U-Net encoder of a latent diffusion model, trained on a denoising-reconstruction objective, learns a representation that is deeply sensitive to the fine-grained details that constitute visual quality---texture fidelity, lighting coherence, sharpness, and subtle color gradients. This "aesthetic prior" appears to be fundamentally more aligned with the FBP task than the class-discriminative features learned by models trained on classification tasks like ImageNet.

\textbf{Second, the necessity of efficient global context modeling.} While the diffusion prior provides a powerful signal for local quality, the ablation results clearly show that it is insufficient on its own. The significant performance leap of the full MD-Net over the Diffusion-only variant underscores the criticality of global facial structure. Facial beauty is not judged on texture alone but on the harmonious interplay of proportions, symmetry, and the geometric arrangement of features. The Vision Mamba stream, with its linear-time complexity and aptitude for modeling long-range dependencies, efficiently captures this holistic context. The synergy is evident: the model learns to evaluate local details within the framework of the global structure.

\textbf{Third, the efficacy of intelligent feature fusion.} The superiority of cross-attention over simple feature concatenation (Table~\ref{tab:ablation_results}, row D) highlights the importance of how the two streams communicate. Cross-attention provides a mechanism for dynamic, context-aware integration. The global representation (from Mamba) can selectively "query" the multi-scale local feature maps (from the diffusion encoder), allowing the model to focus on the most salient aesthetic details given a particular facial structure. This is a more powerful and nuanced fusion strategy than simply combining two independent feature vectors.

Beyond the specific task of FBP, our findings suggest a broader implication for computer vision: complex visual recognition tasks that depend on both micro-level details and macro-level composition may benefit significantly from hybrid architectures that delegate these specialized roles to the most suitable models, such as combining generative and sequential modeling paradigms.

\subsection{Limitations}
\label{sec:limitations}

Despite the strong performance, it is crucial to acknowledge the limitations of this work.

\begin{itemize}
    \item \textbf{Dataset and Annotation Bias:} The foremost limitation is the model's reliance on the FBP5500 dataset. The concept of "beauty" learned by MD-Net is a proxy for the consensus of the 60 annotators for this specific dataset. These annotations are subject to demographic, cultural, and individual biases. Therefore, our model's predictions do not represent an objective or universal measure of beauty but rather reflect the specific distribution and preferences inherent in its training data.
    
    \item \textbf{Interpretability Challenges:} While our dual-stream design is motivated by an intuitive separation of concerns (local vs. global), the model remains a "black box" to a significant extent. The complex, non-linear interactions within the Mamba blocks and the cross-attention module make it difficult to definitively isolate and prove which specific facial attributes (e.g., "symmetry," "skin clarity") are being measured by each component.
    
    \item \textbf{Computational Complexity:} MD-Net is a large and computationally intensive model. Its dual-stream nature, particularly the inclusion of the large U-Net encoder, makes both training and inference more resource-demanding compared to a standard ResNet-50. This presents a practical barrier to deployment in resource-constrained environments.
\end{itemize}

\subsection{Future Work}
\label{sec:future_work}

The success and limitations of MD-Net open up several promising avenues for future research.

\begin{itemize}
    \item \textbf{Fairness and Bias Mitigation:} Directly addressing the dataset bias is a critical next step. Future work should focus on training and evaluating models on more diverse, cross-cultural datasets. Furthermore, employing fairness-aware machine learning techniques to quantify and mitigate biases across different demographic subgroups (e.g., ethnicity, gender) is essential for developing more equitable and responsible models.
    
    \item \textbf{Enhancing Model Interpretability:} To move beyond our architectural intuition, future research could apply advanced explainability techniques. Methods such as Concept Activation Vectors (CAV) could be used to probe whether the Mamba stream is indeed learning concepts like "symmetry," or visualizing the cross-attention maps could reveal which fine-grained features are prioritized for different global structures.
    
    \item \textbf{Model Compression and Efficiency:} To address the computational cost, future work could explore model compression techniques. Knowledge distillation, where the large MD-Net acts as a "teacher" to train a much smaller, faster "student" model (e.g., a MobileNet), could create a lightweight version that retains most of the performance. Quantization and pruning are also viable avenues for creating more efficient versions of the model.
    
    \item \textbf{Generalization to Other Aesthetic Domains:} The core principle of MD-Net---fusing a generative prior for local quality with a sequential model for global structure---is highly generalizable. We plan to investigate the application of this architecture to other subjective, aesthetic assessment tasks, such as predicting the aesthetic quality of photography, art, and user interface design.
\end{itemize}


\section{Conclusion}
\label{sec:conclusion}

In this paper, we addressed the inherent architectural trade-off between local feature extraction and global context modeling in the complex task of Facial Beauty Prediction (FBP). We challenged the prevailing reliance on monolithic architectures by proposing the \textbf{Mamba-Diffusion Network (MD-Net)}, a novel, dual-stream hybrid model designed for synergistic feature representation. Our central thesis was that the nuanced human perception of beauty, which relies on both fine-grained aesthetic details and holistic facial harmony, is best replicated by a system that delegates these specialized roles to the most suitable modern architectures.

Our proposed MD-Net successfully operationalized this principle by integrating two powerful and complementary components:
\begin{enumerate}
    \item A frozen U-Net encoder from a pre-trained latent diffusion model, which provides a rich, unparalleled generative prior for local aesthetic quality.
    \item A fine-tuned Vision Mamba (Vim), a modern state-space model that efficiently captures long-range spatial dependencies and global facial structure with linear-time complexity.
\end{enumerate}
By intelligently fusing these representations using a cross-attention mechanism, MD-Net creates a comprehensive and potent feature space that is simultaneously sensitive to both local texture and global proportionality.

Our comprehensive experiments on the standard SCUT-FBP5500 benchmark have empirically validated our approach. MD-Net not only surpassed strong CNN and Transformer-based baselines but also set a new \textbf{state-of-the-art}, achieving a Pearson Correlation of \textbf{0.9235}, along with the lowest MAE and RMSE scores reported to date. Rigorous ablation studies further confirmed that the synergy between the two streams and the sophisticated fusion mechanism were both critical to this success.

Beyond the specific application of FBP, our work carries broader implications. It demonstrates a promising new architectural paradigm for complex visual assessment tasks: instead of forcing a single architecture to be a generalist, we can achieve superior performance by fusing the specialized strengths of disparate state-of-the-art models, such as those from the generative and sequential modeling domains. We believe that such hybrid, multi-paradigm approaches will be a key driver of future progress in developing more nuanced, human-aligned computer vision systems.


\end{document}